\documentclass[conference]{IEEEtran}
\usepackage{graphicx}
\usepackage{booktabs}
\usepackage[hidelinks]{hyperref}

\title{Tactus: Open-Vocabulary Object Recognition\\from Low-Cost Pressure Arrays}

\author{\IEEEauthorblockN{Abdul Basit Tonmoy}
\IEEEauthorblockA{Eximius Labs \quad Wabash College\\
\texttt{atonmoy27@wabash.edu}\\
\texttt{eximiuslabs.com}}}

\begin{document}
\maketitle

\begin{abstract}
Resistive pressure arrays are the cheapest and most widely shipped tactile sensors, yet
tactile representation learning has concentrated on optical sensors that image a deforming
gel. We present Tactus, an open model that answers text queries from pressure data alone:
on the STAG benchmark (27 objects, held-out recordings), it reaches 0.771 $\pm$ 0.062 top-1
over four runs (top-3 0.935), matching, and at best exceeding, the dataset's supervised
closed-set CNN at 0.76, with no trained classifier head. The recipe is small-data: 187
training recordings, masked-autoencoder pretraining on 144k unlabeled same-sensor frames,
and the sensor's own calibration affine, which recovered more accuracy than every
architecture change combined. The released model's errors concentrate in a few
contact-ambiguous classes, are uncorrelated with text-target geometry (Spearman
$\rho \le 0.05$ over 702 class pairs), and survive paraphrased and even bare-name queries
within one point; two diverse frames recover 89\% of eight-frame accuracy. Failures are
reported with equal precision: cross-sensor pretraining pooling gave no gain, vision
co-training degraded touch, and a mis-normalized input pipeline silently discarded 97\% of
the sensor's dynamic range while producing plausible intermediate results. Weights, code,
and the memory layer the model plugs into are released openly.
\end{abstract}

\section{Introduction}
Touch tells a robot what vision cannot: whether a grasp holds, what is in the hand when the
camera is occluded, when contact slips. The tactile sensors most likely to ship at scale
are low-dimensional resistive and taxel arrays, which cost dollars and already appear in
gloves, insoles, textiles, and robot hands. Why, then, do tactile foundation models read
only optical sensors~\cite{fu2024tvl,higuera2024sparsh,yang2024unitouch,zhao2025t3}, a
signal class requiring camera-grade hardware in every fingertip? The pressure-array side of
the field has strong closed-set classifiers~\cite{sundaram2019stag,lin2024stat} and,
recently, in-the-wild egocentric benchmarks~\cite{song2025opentouch}, but no open model
that connects the signal to language.

Tactus fills that gap. It maps a short window of pressure frames into the frozen text
embedding space of a multimodal model (Qwen3-VL-Embedding-2B via the fusion-embedding
family~\cite{tonmoy2026fusion}), so recognition is cosine ranking against text queries.
Because the same space hosts text, image, video, audio, and inertial motion, touch becomes
one sense inside a cross-modal, language-searchable robot memory rather than an isolated
classifier.

The setting is deliberately hostile to scale: one sensor, 187 training recordings, 27
classes, and a frozen language side that the tactile head must meet where it is. We treat
that hostility as the experimental instrument. With data this small, every recipe decision
is measurable in isolation, and the paper reports the resulting ladder in both directions:
the levers that worked, each quantified with the rest held fixed, and the levers that
failed, each with the mechanism that best explains the failure.

Our contributions:
\begin{itemize}
\item An open pressure-array-to-language model that matches, and at best exceeds, STAG's
supervised closed-set baseline (0.771 $\pm$ 0.062 vs.\ 0.76 top-1) while remaining
open-vocabulary (Table~\ref{tab:results}).
\item A measured small-data recipe: sensor-calibrated normalization, cluster-sampled grasp
windows, and same-sensor masked-autoencoder pretraining, each quantified in isolation
(Table~\ref{tab:ladder}).
\item An error analysis of the released model showing that residual errors are structured
by contact, not by language: confusions concentrate in contact-ambiguous class pairs, are
uncorrelated with text-target similarity ($\rho \le 0.05$, $n{=}702$ pairs), and are
insensitive to query phrasing (Section~\ref{sec:analysis}).
\item Two free inference-time improvements: tuple voting, which averages the embedding
over four independently sampled grasp windows (+4.1 top-1 on the released checkpoint,
Table~\ref{tab:results}), and the finding that two diverse frames already recover 89\% of
eight-frame accuracy (Fig.~\ref{fig:ksweep}).
\item Quantified negative results, including a cross-sensor pooling ablation that
independently reproduces published cross-sensor degradation in a new sensor family
(Section~\ref{sec:negative}), and a forensic account of an input-normalization defect that
silently cost more accuracy than any modeling choice recovered
(Section~\ref{sec:forensics}).
\end{itemize}

\begin{figure}[t]
\centering
\includegraphics[width=\linewidth]{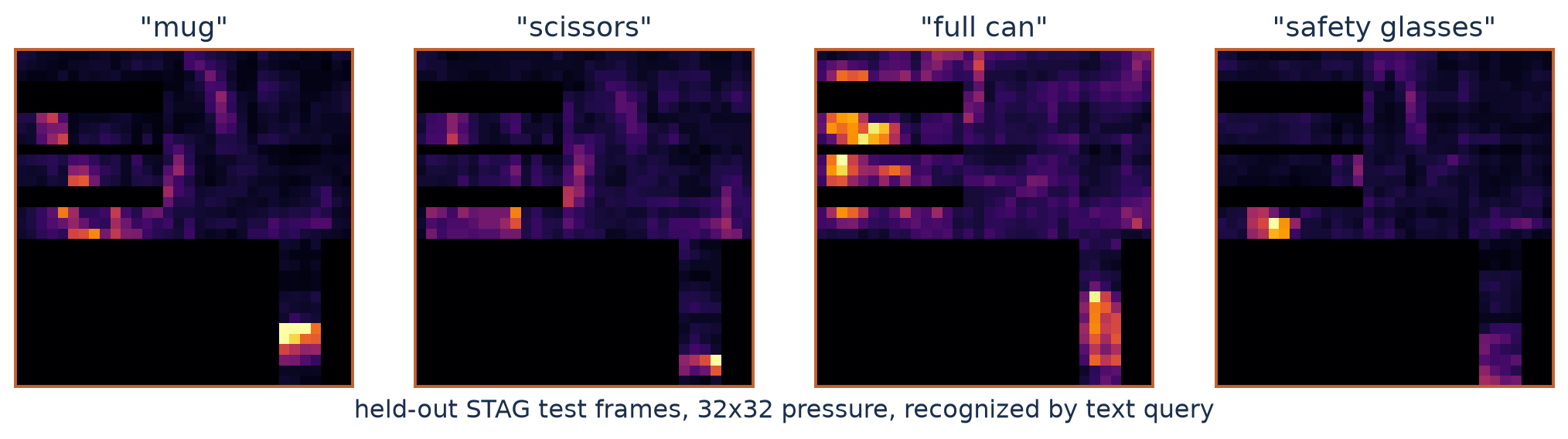}
\caption{Tactus recognizes objects from 32$\times$32 pressure maps alone, by ranking text
queries in a shared multimodal embedding space: no camera, no trained classifier head. The
four frames are real held-out STAG test grasps with their ground-truth query phrases;
each class's most active test frame (largest total pressure) is shown. Across the full
test split the model averages 0.771 top-1 and 0.935 top-3 over 27 such queries
(Table~\ref{tab:results}).}
\label{fig:strip}
\end{figure}

\section{Related Work}
\textbf{Optical tactile representation learning.} TVL~\cite{fu2024tvl} aligns DIGIT images
with vision and language; Sparsh~\cite{higuera2024sparsh} pretrains self-supervised
representations over 460k optical tactile images; UniTouch~\cite{yang2024unitouch} binds
camera-based touch sensors into a pretrained image embedding space with sensor-specific
tokens; T3~\cite{zhao2025t3} spans thirteen sensors with a shared trunk between
sensor-specific encoders and task-specific decoders. All consume RGB images of a deforming
elastomer through vision backbones; none accept the low-dimensional force arrays this paper
targets. Our transfer question is also different in kind: rather than unifying many rich
sensors, we ask how far one cheap sensor can be pushed toward language.

\textbf{Pressure-array recognition.} STAG~\cite{sundaram2019stag} established supervised
object recognition from a 548-taxel glove; published follow-ups reproduce rather than
exceed its accuracy on the same data. STAT~\cite{lin2024stat} adds spatio-temporal
attention with a temporal-order pretraining task for taxel-glove action classification;
TouchFormer~\cite{lyu2026touchformer} fuses taxel force channels for material perception;
conformal tactile textiles~\cite{luo2021textiles} scale the same signal class to garments.
OpenTouch~\cite{song2025opentouch} contributes an in-the-wild egocentric pressure-glove
dataset with a cross-modal retrieval benchmark; it supplies data and evaluation for this
signal class, not a released pressure-to-language model. The classification systems above
are closed-set: recognition requires a classifier trained on the deployment vocabulary.
Tactus keeps the sensor class and removes that constraint.

\textbf{Cross-sensor tactile transfer.} HTT~\cite{bi2026htt} pretrains across
heterogeneous sensors using per-sensor encoders and balanced objectives, and reports that
masked pretraining helps within a sensor while cross-modal alignment can hurt taxel
classification; TacVerse~\cite{wei2026tacverse} measures severe degradation under naive
cross-sensor transfer, even between variants of one sensor. Our pooling ablation
(Section~\ref{sec:negative}) independently reproduces this finding for pressure arrays.

\textbf{Zero-shot recognition against text.} The scoring side of Tactus is CLIP-style
zero-shot classification~\cite{radford2021clip}: class names embedded as natural phrases,
prediction by cosine ranking. Section~\ref{sec:analysis} shows this inheritance is
unusually robust here: paraphrasing, template ensembling, and even bare class names move
accuracy by about one point, because the errors live in the tactile signal rather than in
the text geometry.

\section{The Sensor and the Task}
\label{sec:task}
STAG~\cite{sundaram2019stag} records a 548-taxel resistive glove, rasterized to
32$\times$32 frames at about 30\,Hz, while a wearer manipulates 26 objects plus an
empty-hand class. The classification subset used here contains 135{,}187 frames: 101{,}615
in the train split and 33{,}572 in held-out test recordings. Frames carry three
annotations that are easy to conflate, and the distinction matters in practice:
\texttt{hasValidLabel} marks the 88{,}269 frames (65\%, covering all 27 classes) whose
object label is usable and is the correct grasp filter; \texttt{isGrasp} is a sparse
annotated-instant marker covering only 448 frames, and intersecting the two collapses the
dataset to 77 frames. Early experiments that treated \texttt{isGrasp} as the grasp filter
silently discarded the corpus. The sibling subsets (blindfolded, weights) add 132
recordings of the same 27 objects under different handling conditions; we fold them into
training only, growing the train pool from 55 to 187 recordings, and never into the test
split, which remains the primary subset's held-out recordings throughout.

The task is 27-way open-vocabulary recognition: given a window of pressure frames, rank 27
natural-language grasp phrases (for example ``a human hand firmly grasping a mug'') by
cosine similarity in the frozen text space, with no classifier trained on the label set.
The supervised ceiling on this data is the dataset authors' own CNN at 0.76 top-1, a
closed-set model with a trained decision boundary for every class. Chance is 0.037.

\section{Method}
\begin{figure}[t]
\centering
\includegraphics[width=\linewidth]{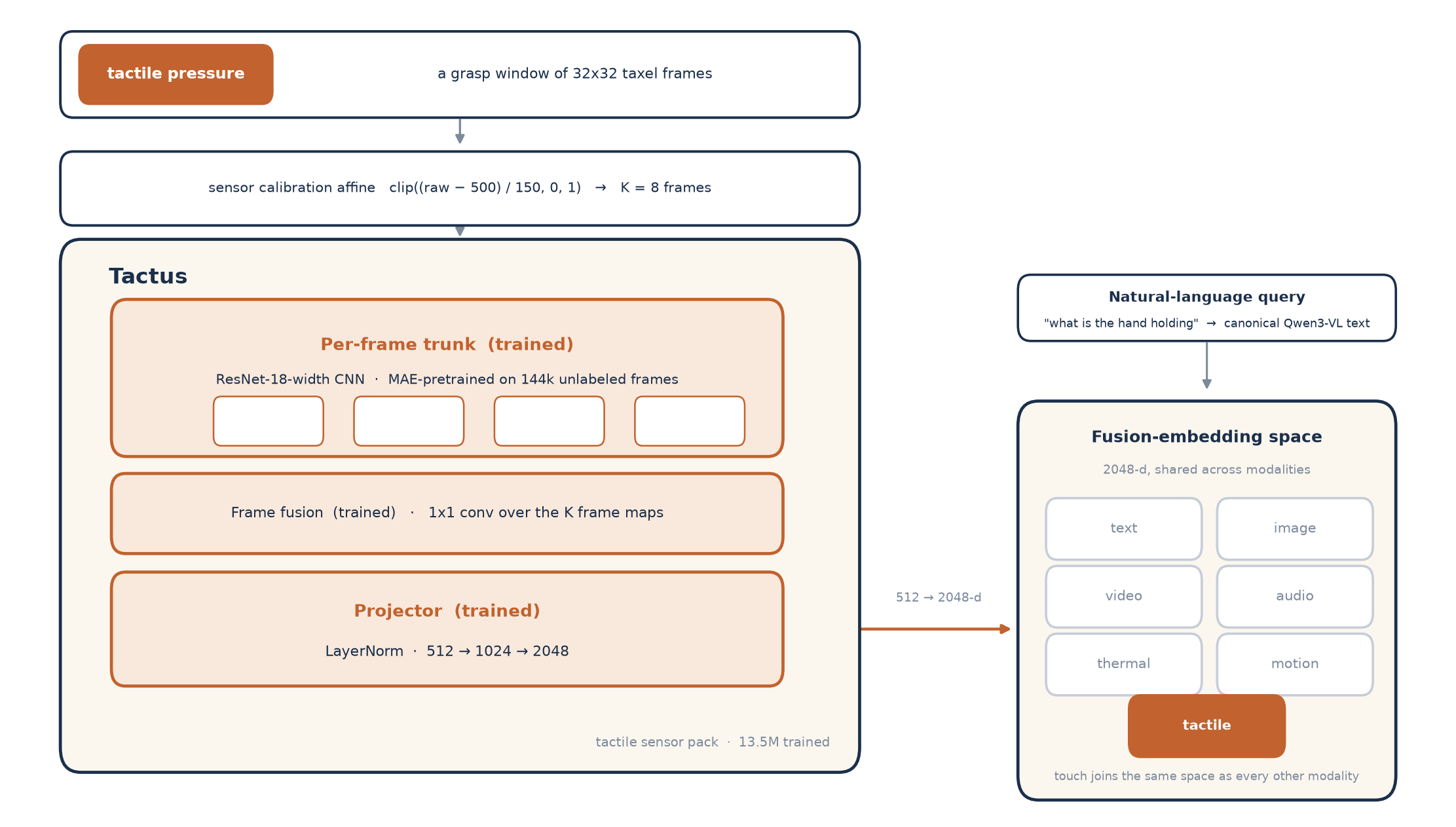}
\caption{Calibrated pressure windows pass through a trained trunk, frame fusion, and
projector into a shared multimodal embedding space; recognition is a natural-language
query against the same space that hosts text, image, video, audio, and motion. Only the
16.2M-parameter head is trained; the language side is frozen, so every existing embedding
in the space is unchanged.}
\label{fig:overview}
\end{figure}

\textbf{Architecture.} Each 32$\times$32 frame passes through a ResNet-18-width trunk
(3$\times$3 stem, four stages); the $K{=}8$ frames of a grasp window are fused by a learned
1$\times$1 convolution over their concatenated feature maps, pooled, and projected
(512$\rightarrow$1024$\rightarrow$2048) into the frozen text space
(Fig.~\ref{fig:overview}). The head totals 16.2M trained parameters (13.5M trunk, 2.6M projector); the language side is
never updated. Nothing about the architecture is exotic; every component was selected by
ablation against the alternatives in Section~\ref{sec:negative}, several of which were
more fashionable and worse.

\textbf{Data path.} Frames are normalized with the sensor's own calibration affine,
$\mathrm{clip}((\mathrm{raw}-500)/150,\,0,\,1)$, following the STAG reference
implementation; Section~\ref{sec:forensics} reports what happened before this was done
correctly. Training windows use STAG's cluster sampling, ported from their
\texttt{ObjectClusterDataset}: each recording's valid frames are reduced by PCA to eight
dimensions, globally min-max scaled, and clustered by $k$-means into $K{=}8$ clusters; each
valid frame becomes the anchor of one training tuple containing one random frame from every
other cluster. Consecutive frames at 30\,Hz are near-duplicates, so an 8-frame contiguous
window carries little more information than one frame; cluster tuples are the same budget
spent on eight genuinely different hand-object configurations. The change raises the
effective training set from 7.7k near-duplicate windows to 74{,}226 diverse tuples,
re-sampled every 500 steps as in the original.

\textbf{Text targets.} Each class is embedded as a natural grasp phrase through the
family's canonical text readout and L2-normalized. The 27 phrases share most of their
wording, so their embeddings carry a large common component: mean off-diagonal cosine
0.504, nearest-distractor margin 0.354. Removing the class centroid and re-normalizing
spreads the targets to off-diagonal mean $-0.038$ (the maximum for 27 unit vectors) and
margin 0.777. Training is label cross-entropy against these targets at temperature 0.07
(16k steps, AdamW, cosine schedule with warmup, spatial and temporal augmentation).

\textbf{Pretraining.} The trunk is initialized by masked-autoencoder
pretraining~\cite{he2022mae} on 144k same-sensor frames, including the 47k frames whose
labels are unusable for supervision, with mask ratio 0.6 (the taxel setting of
HTT~\cite{bi2026htt}) and per-patch normalized targets; supervised test frames are excluded
from the pool. Per-patch normalization also buys a free diagnostic: a mean-predicting
encoder is pinned at loss 1.0, so reconstruction-objective collapse is visible on the loss
curve rather than discovered downstream. Pretraining costs about 65 minutes on one A10G.

\textbf{Evaluation.} Test recordings are tiled into cluster tuples with a fixed seed; each
tuple is scored by cosine ranking against the 27 class phrases. We report tuple-level
top-1/top-3 and a recording-level score pooling up to 64 frames per test recording. Means
aggregate independent runs with identical hyperparameters. One caveat travels with every
number: cluster sampling changes the test population as well as the training set, so
tuple-level accuracies are comparable only within one protocol. The recording-level score,
which pools a recording's valid frames without any sampling, is the protocol-independent
anchor; it is also the number that reproduced to four decimal places when we replayed the
released checkpoint in a fresh environment (Section~\ref{sec:analysis}).

\section{Main Results}
Table~\ref{tab:results} reports 27-way recognition on STAG's held-out test recordings,
scored open-vocabulary. The recipe mean exceeds the original supervised closed-set CNN,
though by less than one standard error; we therefore characterize the result as matching,
and at best exceeding, the original baseline while performing a harder task. Top-3 is
0.935 on the mean and 0.951 on the released checkpoint. Our evaluation mirrors STAG's
cluster-sampling test protocol but is not the authors' byte-identical harness.

\begin{table}[h]
\centering
\caption{STAG classification, 27 objects, held-out test recordings. Top-1 means carry
$\pm$ one sample standard deviation over independent runs; other columns are means over
the same runs. Bold marks the primary result; the indented rows are one of the four
recipe runs (the released checkpoint), evaluated with a single grasp-window draw and with
tuple voting over four draws.}
\label{tab:results}
\begin{tabular}{lccc}
\toprule
 & top-1 & top-3 & recording top-1 \\
\midrule
Chance & 0.037 & 0.111 & 0.037 \\
STAG supervised CNN~\cite{sundaram2019stag} & 0.76 & -- & -- \\
Scratch (no MAE), mean of 3 & 0.705 $\pm$ 0.033 & 0.905 & 0.691 \\
Tactus, mean of 4 & \textbf{0.771 $\pm$ 0.062} & \textbf{0.935} & \textbf{0.722} \\
\quad released checkpoint (single run) & 0.817 & 0.951 & 0.741 \\
\quad + tuple voting over 4 draws & 0.858 & 0.970 & 0.741 \\
\bottomrule
\end{tabular}
\end{table}

Table~\ref{tab:ladder} traces the recipe: each row is one measured change with the rest
held fixed. Data-path corrections contribute more than every architecture change combined;
the largest single step is repairing the input normalization, the second largest is
same-sensor pretraining. Everything above the rule in Table~\ref{tab:ladder} was measured
on 3\% of the sensor's dynamic range; Section~\ref{sec:forensics} is the account of
finding that out.

\begin{table}[h]
\centering
\caption{Recipe ladder (27-way top-1). Rows above the rule predate the normalization
correction and are shown for trajectory; their absolute values are depressed by defective
input scaling.}
\label{tab:ladder}
\begin{tabular}{lc}
\toprule
Configuration & top-1 \\
\midrule
Single frames, no valid-frame filter & 0.194 \\
+ grasp windows, valid-frame filter & 0.299 \\
+ cluster sampling, conv fusion, ResNet trunk & 0.377 \\
+ cosine schedule & 0.396 \\
+ ResNet-18 width $\times$ 16k steps & 0.468 \\
+ augmentation & 0.574 \\
\midrule
Calibration affine restored (scratch, mean of 3) & 0.705 \\
+ same-sensor MAE initialization (mean of 4) & \textbf{0.771} \\
\bottomrule
\end{tabular}
\end{table}

\section{What the Released Model Gets Wrong, and Why}
\label{sec:analysis}
Headline accuracies say little about structure, so we re-scored the released checkpoint
(the exact artifact on the model hub, at its release tag) over all 26{,}250 test tuples and
dumped the full similarity rows. Replaying the recording-level protocol reproduced the
release evaluation to four decimal places (0.7407), confirming artifact and pipeline
integrity; the analyses below share one environment and protocol (tuple voting over four
draws, template-ensembled targets) and are internally comparable.

\begin{figure}[t]
\centering
\includegraphics[width=\linewidth]{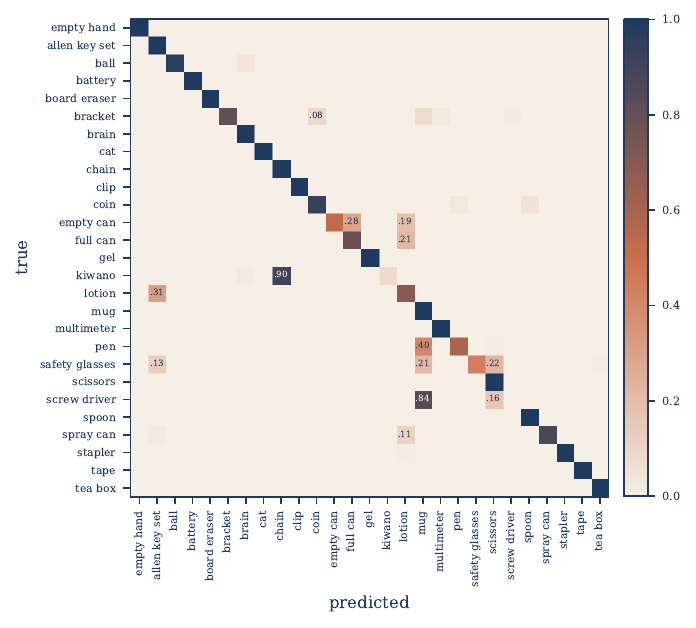}
\caption{Row-normalized confusion matrix of the released checkpoint over all 26{,}250 test
tuples. Errors are sparse and structured: most classes sit on the diagonal at 0.80 or
higher, and the mass off the diagonal concentrates in contact-ambiguous pairs
(kiwano$\rightarrow$chain, screw\_driver$\rightarrow$mug), not in linguistically similar
ones. Cell labels mark off-diagonal rates $\ge$ 0.08.}
\label{fig:confusion}
\end{figure}

\textbf{Errors concentrate, and they concentrate by contact.} Twenty of 27 classes score at
or above 0.80 top-1, sixteen at 0.99 or higher. The overall error is dominated by two
classes: \texttt{kiwano} (0.08 top-1) and \texttt{screw\_driver} (0.001). Both failures are
legible in contact terms. The kiwano, a spiked fruit, presses into the palm as a
constellation of discrete points, and 90\% of its tuples are predicted as \texttt{chain},
the other object in the vocabulary that contacts as many separated points; the model's
second choices are good enough that kiwano's top-3 is 0.933. The screwdriver is a thin
cylindrical grip predicted as \texttt{mug} (84\% of tuples), the vocabulary's other
cylindrical handle. The remaining notable confusions follow the same pattern:
\texttt{pen}$\rightarrow$\texttt{mug} (0.40), \texttt{empty\_can}$\rightarrow$
\texttt{full\_can} (0.28), \texttt{safety\_glasses}$\rightarrow$\texttt{scissors} (0.22,
two thin rigid arms). Fig.~\ref{fig:confusion} shows the full matrix.

\textbf{The text space is not the error source.} A natural worry for open-vocabulary
scoring is that errors are inherited from the text encoder: classes whose phrases embed
similarly should confuse. We test this directly. For every ordered class pair we compare
the confusion rate against the cosine similarity of the class text targets, before and
after centering. The rank correlation is negligible: Spearman $\rho = 0.024$ (uncentered)
and $0.053$ (centered) over 702 pairs. The two catastrophic confusions have unremarkable
text cosines (kiwano/chain 0.43, screw\_driver/mug 0.47), while the vocabulary's most
similar pair by text (\texttt{empty\_can}/\texttt{full\_can}, cosine 0.88, the same object
differing only in fill state) confuses at just 0.28. Whatever the model still gets wrong,
it is wrong about pressure, not about language.

\textbf{Query phrasing barely matters.} Scoring the same embeddings against six target
constructions (template ensemble, single template, bare class names; each centered and
uncentered) spans 0.850 to 0.860 top-1. Even bare names (``mug'') match the hand-tuned
ensemble. Combined with the correlation result above, this closes the question from both
directions: the text side neither causes the errors nor requires prompt engineering to
avoid them. We attribute this robustness to target centering during training, which removes
the shared-wording component that phrasing choices mostly perturb.

\begin{figure}[t]
\centering
\includegraphics[width=0.85\linewidth]{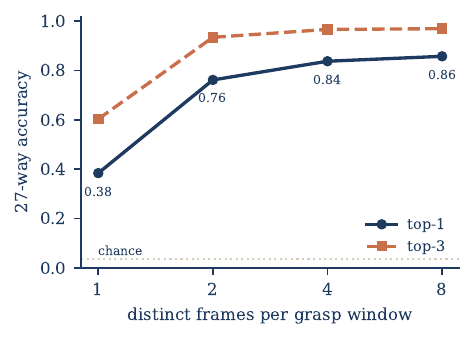}
\caption{Accuracy of the released checkpoint vs.\ distinct frames per grasp window
(inference-time only; the trained fusion is unchanged). One frame is not enough (0.38);
two recover 89\% of the eight-frame accuracy. Touch is temporal, but shallowly so.}
\label{fig:ksweep}
\end{figure}

\textbf{Touch needs integration, but shallow integration.} Fig.~\ref{fig:ksweep} sweeps
the number of distinct frames per window at inference (repeating the last frame to keep
the trained 8-frame fusion intact). A single cluster-diverse frame collapses to 0.38
top-1; two frames recover 0.762, which is 89\% of the 8-frame 0.858; four reach 0.84. A
grasp is not legible from one still pressure image, but two genuinely different
hand-object configurations nearly saturate the task. For deployment this sets the latency
budget: at 30\,Hz, two diverse contacts arrive within a fraction of a second of
manipulation.

\textbf{Tuple voting is a free four points.} The released evaluation scores one cluster
tuple per test anchor. Averaging the embedding over four independent tuple draws before
ranking (test-time augmentation over the sampling noise, not over the data) lifts the
released checkpoint from 0.817 to 0.858 tuple-level top-1 at four times the inference
cost and no retraining. Recording-level accuracy is unchanged, as expected: the recording
protocol already pools across the recording and has no sampling noise to vote away.

\textbf{Pooling changes the failure surface rather than shrinking it.} At the recording
level (all of a recording's valid frames pooled into one embedding) the released model
resolves 20 of 27 test recordings. Three tuple-level failure classes persist
(kiwano$\rightarrow$chain, empty\_can, pen), but the extremes do not transfer:
\texttt{screw\_driver}, at 0.001 tuple-level, resolves correctly when its recording is
pooled, while \texttt{clip}, at 1.000 tuple-level, misses. The pooled window spans
pre-grasp and transition frames that the cluster tuples exclude, so it is a different
input distribution, not a vote over the same one. With one recording per class, each
recording-level verdict is a single sample; we read the tuple-level matrix as the
reliable error structure and the recording score as the deployment-facing aggregate.

\section{What Did Not Work}
\label{sec:negative}
Negative results shaped this recipe more than positive ones; the small-data tactile
setting makes several intuitive choices fail, and each failure below comes with its most
probable mechanism and the evidence for it.

\textbf{Cross-sensor pretraining pooling gives nothing.} Pretraining on a pool dominated
93/7 by a foreign 16$\times$16 taxel corpus produced zero downstream gain over training
from scratch (0.68 vs.\ the 0.71 scratch band), even with equal-per-source batch balancing
and normalized reconstruction targets, while same-sensor pretraining, with thirteen times
fewer frames than the foreign corpus, was worth +6.6 points on the mean (0.705
$\rightarrow$ 0.771, Table~\ref{tab:ladder}). This independently reproduces, in a new sensor family,
the cross-sensor degradation reported by HTT~\cite{bi2026htt} and
TacVerse~\cite{wei2026tacverse}: the trunk's features organize around the dominant sensor's
statistics, which do not transfer. Under proportional sampling the foreign corpus
additionally masked the reconstruction objective: the loss settled at the data variance,
the signature of a mean-predicting solution, which per-patch target normalization makes
visible by pinning that solution at loss 1.0. The practical rule we now follow: a
same-sensor-only control is mandatory before trusting any pooled pretraining number.

\textbf{Vision co-training hurts touch.} Adding a contrastive touch-to-image objective
against synchronized camera frames gave nothing at weight 0.5 and degraded tactile
recognition at weight 1.0 (0.39 and 0.28, against a 0.39 text-only control at the time of
the ablation), despite the same lever having improved our inertial-motion pack.
The mechanism we propose: the paired images are dominated by hand and scene appearance
rather than object identity, so the alignment target injects nuisance structure. The
contrast with UniTouch~\cite{yang2024unitouch}, where vision alignment is the central
lever for optical sensors, is informative: an optical tactile image shares appearance
statistics with photographs, and a pressure map does not.

\textbf{The text targets are not the bottleneck.} A learnable 27-way linear classifier
head, free to place its own decision boundaries (STAG's own setting run on our encoder),
scored below the frozen-text-target head at the time of the diagnostic (0.33 vs.\ 0.396).
If the fixed targets were costing accuracy, the free head would have recovered it. This
diagnostic, together with the $\rho \le 0.05$ result of Section~\ref{sec:analysis},
locates the remaining headroom in the tactile encoder, not the language side.

\textbf{Capacity without regularization collapses.} A from-scratch temporal-attention
encoder never left chance on 55 recordings (training loss flat at $\ln 27$); STAT's
success with attention on taxel data~\cite{lin2024stat} comes with a dedicated temporal
pretraining task, which supports the reading that attention needs an auxiliary objective at
this scale rather than being unusable. The 13.5M-trunk model overfit catastrophically when trained
past its optimum without augmentation. A frame-differencing motion channel hurt under
cluster sampling for a structural reason: tuple members are drawn from different clusters,
so adjacent slots are not adjacent in time and the difference channel carries noise
rather than motion. Augmentation itself reversed verdict with scale: on a 0.9M-param
model it was destabilizing (repeats spanning 0.15 to 0.48 under one configuration), while
on the 13.5M-trunk model it contributed over +10 points (0.47 to 0.57). Each of these verdicts
reversed under a scale change, which cautions against fixed conclusions from
single-configuration ablations; the augmentation case also taught us to run full repeat
sets before declaring a lever dead, since its first two repeats looked like noise around
zero and its fourth was the best number we had seen to that point.

\section{Anatomy of a Silent Data Defect}
\label{sec:forensics}
The largest error source in this project was not a modeling choice. We document it in
detail because nothing about it was visible in the places one normally looks, and because
the fix outweighed every architectural contribution combined.

Our initial ingest normalized each frame by the corpus maximum. The STAG sensor, like most
resistive arrays, has a resting pedestal: untouched taxels read approximately 500 raw, not
zero, and the informative band ends near 650. Corpus-max scaling preserved the pedestal, so
cached frames sat near 130 of 255 with a 95th percentile of 132: roughly 8 usable intensity
levels, 3\% of the encoded range. Measured raw ranges across STAG's subsets (including
handposes, which enters only the pretraining pool) were 446 to 1000 (classification), 509
to 991 (blindfolded), 510 to 992 (handposes), and 510 to 654 (weights); the weights subset never saturates, so its per-subset maximum landed it on a
1.53$\times$ different scale from its siblings, a second defect hiding inside the first.

Three properties made this defect expensive. It was invisible in loss curves: models
trained, losses fell, and intermediate accuracies looked like publishable ablations (the
entire upper half of Table~\ref{tab:ladder} was measured on 3\% contrast). It was
deterministic: adding the mis-scaled sibling data produced training collapses to 0.128
that reproduced to four decimal places across repeats, which we initially attributed to
optimization and attacked with gradient clipping and temperature schedules; the optimizer
was never the problem. And it inverted a conclusion: the sibling subsets looked useless or
harmful under the defect and are worth several points after the fix.

The repair is the sensor's own calibration, $\mathrm{clip}((\mathrm{raw}-500)/150, 0, 1)$,
applied identically to every subset. After it, cached frames span 30 to 60 usable levels,
the cross-subset scale ratio is 0.944, and the collapse basin is gone (zero collapses in
all post-fix runs). Two guards now stand: a dynamic-range regression gate in the ingest
diagnostics, and the rule that every pretraining pool change reruns the same-sensor
control. We recommend both to anyone training on resistive arrays; the failure mode is
generic to any sensor with a pedestal, and per-corpus max normalization is the obvious
first thing to write.

\section{Discussion}
Three observations generalize beyond this dataset. First, in small-data tactile learning
the data path dominates: sensor calibration, frame selection, and label hygiene moved
accuracy far more than architecture, and the single largest gain in the paper is a
two-constant affine from the sensor's datasheet. Second, self-supervision pays where
labels are scarce but frames are not: a third of our corpus was unusable for supervision
yet contributed through reconstruction pretraining. Third, transfer across tactile sensor
families remains unsolved for force arrays just as for optical sensors; treating each
sensor family as its own pretraining domain is currently the only recipe with evidence
behind it.

The error analysis adds a fourth observation specific to open-vocabulary sensing. The
worry that zero-shot scoring imports the text encoder's failure modes is, here,
unfounded: errors are contact-structured, query-phrasing-insensitive, and uncorrelated
with text geometry. The practical reading is optimistic for the recipe: a frozen language
side is not a tax on a well-trained sensor head, and the remaining errors point at the
sensor and the data (kiwano and chain genuinely feel alike at 32$\times$32) rather than at
the scoring mechanism. Where the vocabulary contains a truly contact-identical pair
(empty versus full can), the model resolves it imperfectly but far above its text-side
confusability, suggesting weight and stiffness cues survive the rasterization.

Within the fusion-embedding family, Tactus also fixes one point on a design map. The
family supports two routes for adding a sense to a frozen base: in-layer gated capacity
inside the language model, used when the new signal must traverse the frozen stack itself,
and an external encoder-plus-projector head, used when a small trained encoder suffices.
Pressure arrays, like inertial motion and unlike audio and thermal imagery, take the
second route: 16.2M trained parameters against a byte-frozen base, with every existing
embedding unchanged. The model plugs into the same released memory layer as the other
senses, so eval traces and robot session logs gain touch as a queryable channel with no
change to the rest of the stack.

If one rule carries over to other low-cost-sensor efforts, it is this: calibrate and
diversify the input before touching the model, and keep a same-sensor control running at
all times.

\section{Limitations and Release}
Run-to-run variance ($\pm$0.062 top-1 across four seeds, spanning 0.701 to 0.829) is the
main open engineering problem; the released checkpoint is a single run from that set, and
we report the mean alongside it. Results cover one sensor family; our own pooling ablation predicts
that transfer to other taxel geometries requires fine-tuning. Open-vocabulary here means
text queries over the evaluation categories, not validated open-set generalization to
novel object classes. The analysis of Section~\ref{sec:analysis} uses a slightly stronger
inference protocol (tuple voting, template ensembling) than the release evaluation; both
are reported in Table~\ref{tab:results} and the recording-level anchor is identical under
both. Weights (CC-BY-NC-4.0, following the training data's license), inference code, and
the Apache-2.0 session-memory layer the model plugs into
(\texttt{pip install engram-robomem}) are available via
\texttt{huggingface.co/EximiusLabs}.

\end{document}